\pdfoutput=1
\documentclass[11pt]{article}

\usepackage{acl}

\usepackage{times}
\usepackage{latexsym}
\usepackage{graphicx}

\usepackage[T1]{fontenc}
\usepackage[utf8]{inputenc}

\usepackage{microtype}

\usepackage{inconsolata}

\newcommand{\mybox}[2]{\noindent\framebox{\parbox{\dimexpr1.0\linewidth-2\fboxsep-2\fboxrule}{%\itshape%
  \textbf{#1}\\#2}}}
\newcommand{\ibd}[0]{\textit{ibd.}}

\title{Challenges and Opportunities of NLP for HR Applications: \\ A Discussion Paper\thanks{The authors acknowledge the support of the Hightech Agenda Bavaria R\&D funding program of the State of Bavaria to the first author. We would like to thank Hedwig Schmid and Tim Nugent and three anonymous reviewers for discussions.}

\author{Jochen L.~Leidner${}^{1,2,3}$ \and Mark Stevenson${}^{3}$ \\[2mm]
  ${}^{1}$ Coburg University of  Applied Sciences, Friedrich-Streib-Straße 2, 96450 Coburg, DE \\
  ${}^{2}$ %KnowledgeSpaces\textsuperscript{\textregistered} UG\,(haftungsbeschränkt),
  Erfurter Straße 25a, 96450 Coburg, DE \\
  ${}^{3}$ University of Sheffield,  Regents Court, 211 Portobello, Sheffield S1 4DP, UK \\
  \texttt{leidner@acm.org} / \texttt{mark.stevenson@sheffield.ac.uk} \\}
}

\begin{document}

\maketitle

% ----------------------------------------------------------------------

\begin{abstract}
  Over the course of the recent decade, tremendous progress has been made in the areas of machine learning and natural language processing, which opened up vast areas of potential application use cases, including hiring and human resource management.

  We review the use cases for text analytics in the realm
  of human resources/personnel management, including actually
  realized as well as potential
  but not yet implemented ones,
  and we analyze the opportunities
  and risks of these.

\end{abstract}

% ----------------------------------------------------------------------

\section{Introduction}

Over the last decade, Artificial Intelligence (AI), specifically Natural Language Processing (NLP), driven by advances in
Machine Learning (ML), has made substantial
progress to the level that now many
applications are now enabled, including but
not limited to the realms of hiring and Human Resource Management (HRM).
Furthermore, the SARS CoViD-19 pandemic
has led to a form of ``forced radical digitalization'' of the workplace (then the home office), and working from home for knowledge and office workers is now
much more broadly accepted than before.

\begin{figure*}[!ht]
    \centering
    \includegraphics[width=\textwidth]{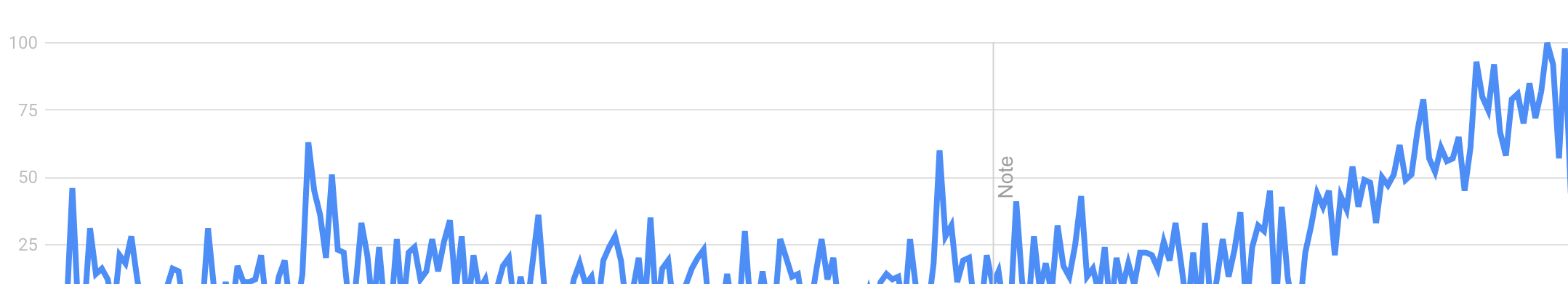}
    \caption{Google Trends: Joint Use of ``HR'' and ``AI'' in Web Search Queries over Time (2018-2023)}
    \label{fig:trends}
\end{figure*}

Accordingly, in business practice and
economic research the interest in the
``HRM analytics'' space has been rising: 
recently, Web search for
``HR and AI'' have been increasing in frequency (Figure \ref{fig:trends}).

However, some high-profile early adopters
of automation support in HRM have also
encountered high-profile failures
\cite{Dastin:2018}.

% At first, the list of application use cases for NLP sounds potentially very appealing:
HR provides multiple potentially appealing use cases for NLP, including: 
\begin{itemize}
\item interactive employee self-service chatbots \cite{Aqel-Vadera:2010:ISWSA};
\item generation of personalized HR documentation (e.g. standardized appraisal letters, Frequently Asked Questions (FAQ) collections);
\item automatic mining of HR-relevant intelligence (demand and supply, availability/change readiness and salary statistics);
\item hiring assistant: help to formulate/generate \cite{Indeed:2023:online}, review, predict the effectiveness of job postings \cite{LinkedIn:2023:online,Qin-etal:2023:TKDE}; filtering/ranking of incoming applications, information extraction of contact information, skills, experiences, and biographic details from resumes \cite{Khaouja-Kassou-Ghogho:2021:IEEEAccess,Zhang-etal:2022:NAACL});
\item matching engines/recommendation systems that match job seekers with employers \cite{Yang-etal:2022:RecSys};
\item develop missing skills to get ready for a role or career \cite{JoseGarcia-etal:2022:IJAIEd};
\item retention support (attrition modeling/predictive analytics);
\item performance monitoring support and appraisal assistance \cite{Brindha-Santhi-2012-ICMIEE};
\item business intelligence (e.g. open source competitive intelligence via analysis of company job postings to infer competitors' plans);
\item automated or computer-supported on-boarding or leaving assistance.
\end{itemize}
An employee self-service chatbot can facilitate access to answers for employee questions, but it may
also decrease the sense of belonging and weaken the human bond between employee and HR representative\footnote{One employee interviewed on the topic stated ``It used to be that the personnel department was a person we knew that cared about how we are; now, it feels HR is on the corporation's side.'' Another (European) employee stated ``Recently, HR has become an anonymous e-mail alias that points to a function outsourced to India.''}.
%
%% Both in the head of a potential employee and in the head of a hiring manager representing an employing candidate, there are two workflows that the respective parties follow, implicitly or consciously (Figure \ref{fig:workflow}).

This position paper reviews existing work in order to get a sense of
the state of the art in the field and then identifies and analyses issues
that are either immanent in the application of NLP techniques to
human resources in general, or issues that may materialize, depending
on the occasion and situational context. Opportunities
arising from this space are also described.
This paper focuses on the following research questions: 
% A number of research questions are of interest when analyzing the space of ``NLP for HR'' or ``HRM text analytrics'', and we can only begin to delve into these:
\begin{description}
\item[RQ-1]  What types of support can NLP enable within HR? 
\item[RQ-2]  How should the support differ from the employer's side versus the employee's side? 
\item[RQ-3]  What opportunities versus challenges/risks result from the recent technological advances? 
%\item \todox{ more? }
\end{description}

We believe that while there are many
application papers describing the technical
side of application of NLP to HR, and
many business papers analyzing the business side of HR text analytics and its
impact on the business once it is introduced,
we are the first to apply a systematic
\textbf{employment lifecycle} framework by
asking at each stage before or during the
employment journey what applications of NLP
(in the broadest sense, including any ML over words) does now, or can potentially in the
future, support said process/workflow step.

% ----------------------------------------------------------------------

\section{Related Work}

\begin{table*}[t]
    \centering
\mybox{Computer-Aided HRM Glossary}{\
• Automatic Resume/CV Mining (AMM): to proactively source (collect, crawl) resumes from the open Web \\ \
• Automatic Resume/CV Screening (ARS): to parse/analyze a given resume to extract contact details, KSAOs, past positions held etc. \\ \
• Applicant Tracking System (ATS): a (software) system for executing a prescribed hiring workflow for each applicant, defined by a set of stages and associated actions. \\ \
• Business Intelligence (BI): obtaining actionable intelligence about a business, including especially one's own business; often meant to include automatic support via BI systems, machine learning (ML) and Big Data (BD) approaches. \\ \
• Computer-Aided Human Resource Management (CAHRM): the intersection of applied computer science and personnel management (NLP/ML meets HRM) \\ \
• Competitive Intelligence (CI): obtaining actionable information about a business' direct competitors (e.g. through their hiring plans). \\ \
• Continued Professional Development (CPD): the ongoing process of developing, (re-)educating and upskilling one's workforce, as it is required by many professional bodies (e.g. PMI) \\ \
• Employee Turnover Prediction (ETP): the task of early identification of likely leavers, similar to churn detection in CRM. \\ \
• Knowledge, Skills, Abilities, and other Occupation-Related Characteristics needed for a job (KSAOs). }
    \caption{Glossary of Terms}
    \label{tab:glossary}
\end{table*}

\textbf{NLP Applied to HR(M).}
Writing good job postings is important to signal to potential applicants the  Knowledge, Skills, Abilities, and other Occupation-Related Characteristics needed for a job (KSAOs).\footnote{See Table \ref{tab:glossary} for expansions of acronyms used here.} \citet{Mujtaba-Mahapatra:2020:CSCI} and \citet{Putka-etal:2023:JBusPsy} use an occupational taxonomy and NLP tools to mine KSAOs from online job postings. The former apply the syntactic parser from SpaCy and GloVe embeddings are used to perform key-phrase extraction, and the latter rely on a combination of TreeTagger and the R library koRpus to the same end.
\citet{Urbano-etal:2021:LDK} describe
a method to detect contradictions and
ambiguities in job ads before they are
posted.
\citet{Varelas-etal:2022:IFIPWS} describe a method for the automatic classification of free-text Web job vacancies against the standard ISCO taxonomy of occupations, which is helpful for occupation macro-statistics.

Information Extraction (IE) from resumes
is perhaps the NLP task most commonly
associated with the domain of hiring and HR;
for instance, \citet{Vukadin-etal:2021:IEEEAccess}
present a system for data extraction from
resumes in multiple languages 
Special companies -- such as
Textkernel BV\footnote{\url{https://www.textkernel.com} (accessed 2023-12-08)} and Daxtra Ltd.\footnote{\url{https://www.daxtra.com} (accessed 2023-12-08)}
-- exist that focus entirely on this topic,
and technologies to automate it have been
integrated in some mainstream HR software
support systems.
\newcite{Yu-etal-2005:ACL} compare SVMs and HMMs on Chinese resume IE and
find that SVMs (77\%) outperforms HMMs (90\% or higher token overlap were defined as matches).
Personal information is reportedly easier to extract than educational information.
\citet{Li-etal:2021:ICCT} as well as \citet{Gan-Mori:2023:NLDB} present more recent, neural transformer based methods
for resume extraction, whereas \citet{Liu-etal:2020:ACL} propose the opposite task of job posting \emph{generation}. In particular,
\citeauthor{Gan-Mori:2023:NLDB}'s work compared Masked Language Models (MLM) and sequence-to-sequence Pretrained Language Models (PLMs) regarding their ability to extract resume information with prompting in a various few-shot scenarios.

\citet{Kadam-etal:2022} present a system that assists the screeners in effectively shortlisting the resumes by automatically
extracting candidate’s abilities, job experience, designation, and degree, and by using this information
to match candidates against role profiles.
% Challenge: selecting the best candidate for the enrollment cycle from a large pool of candidates has been a significant challenge
% Solution: automatic matching to create a shortlist, human selection

\citet{Weaver:2017:PhD} presents a Ph.D. thesis concerning the prediction of employee
performance based on resumes/biodata using the lexical resource LIWC for sentiment analysis.

%\textbf{Churn Prevention \& Prediction.}
\citet{Yordanova-Kabakchieva:2022:AIPC}
use $K$-means clustering of resumes to group together likely loyal employees.
\citet{Punnoose-Ajit:2016:IJAdvResAI} predict
employee turnover in a large retail company
using XGBoost and other methods using 33 features (AUC=0.86).
Whereas the former work is proactive, the
latter is predictive.
\citet{Saradhi-Palshikar:2011:ExpSysApp}
and 
\citet{Punnoose-Ajit:2016:IJAdvResAI}
also predicted employee turnover, using
SVM and XGBoost, respectively.

\citet{Guo-etal:2020:HRMJ} demonstrate two case studies, a first one applies sentiment analysis to comments about companies to score employee popularity,
which is important for selecting a potential employer, and a second application  demonstrates how candidate employee personality (self-reported scores on the Big Five) is revealed by applying doc2vec to situational interview questions.

\citet{Zhu-Hudelot:2022:NAACL}
mine resumes for job titles and associated needed skill-sets in order to obtain
role promotion paths with the ultimate aim at generating better role titles. They construct
graphs from the extracted material of the form:
purchase agent $\rightarrow$ purchasing manager $\rightarrow$ staff account purchasing manager (\ibd{}, p. 2135).

In their meta-analysis \citet{Tambe:2019:CalifMgtRev} acknowledge
the existing gap between the promise of
AI on the one hand and the current state of
real HRM on the other hand.
They then proceed to identify four challenges in using AI techniques for HR tasks, namely 1. complexity of HR phenomena, 2. constraints imposed by small data sets, 3. accountability questions associated with fairness and other ethical and legal constraints, and 4. possible adverse employee reactions to management decisions via data-based algorithms.
Interestingly, they also propose to address these by
identifying practical responses based on three overlapping principles: 1. causal reasoning, 2. randomization and experiments, and 3. employee contribution, and they claim
following them was economically efficient and
socially appropriate.

\citet{JoseGarcia-etal:2022:IJAIEd}
develop a Career Coach working from a given set of skills and a stated desired career profile, a set of skill gaps are
derived as the difference set, and additional training is recommended to fill these; obviously, this is an idea that can be used 
for getting unemployed people back into
the workforce as well as offering it to
employees in a job to develop for future
positions.

There is no shortage of (mostly very recent)
articles on the topic in the business world.
\citet{Campion-etal:2020} is a review of 341 articles form the management literature that use, review, or promote text analytics as applied to HR.

\citet{Bruera-etal:2022:ELIHLTWS} describe
experiments to generate synthetic resumes,
which is a helpful sub-task in NLP for HR research in jurisdictions such as the European Union, where personal data is
especially protected; as resume analyzers
under development will need a lot of data
for development, training and/or testing,
automatically generated resumes for non-existing people offer a viable path.\footnote{\cite{Silva-etal:2020:INFOCOMWS} discuss the alternative path of anonymizing resumes of real
people.}
Not all job postings circulated in production systems are genuine either prompting researchers such as 
\citet{Mahbub-Pardede:2018:ISD},
\citet{Amaar-etal:2022:NeurProcLett} to explore automated detection of scam job postings.

\textbf{Ethics \& NLP.}
\citet{Leidner-Plachouras:2017:EthicsNLPWS}
present a range of early
examples of ethical violations
in natural language processing
research, development and system deployment, and they propose
a structured process with
ethics checkpoints to mitigate
risk.

\textbf{Fairness \& ML.}
\citet{Hutchinson-Mitchell:2019:FAT} survey
five decades of approaching fairness in
machine learning, 1960-2010; their Table 2 (\ibd{}, p. 53) shows 16 mathematical formalizations of
fairness. The annual conference \textit{Fairness Accountability and Transparency {FAT$\ast$, formerly FATML}}, at which the paper by \citeauthor{Hutchinson-Mitchell:2019:FAT} appeared, is a good
venue to find more works from this space.
\citet{Eitle-etal:2021:ECIS} are concerned
with the question of fairness arising from
the use of job recommendation engines in HR.

In the context of the unmet need for
healthcare personnel,
\citet{Mesko-etal:2018:BMCHealthServRes}
argue that while AI is not meant to replace caregivers, those who use AI will probably replace those who do not; they discuss
ethical questions in that particular domain.

% \todox{ review the remaining literature }
%\citet{Tonneau-etal:2022:ACL}
%\citet{Li-etal:2020:EMNLP}
%\citet{Liu-etal:2016:ACL}
%\citet{Delecraz-Eltarr-Oullier:2022:ELHLTWS}
%\citet{Hemamou-Coleman:2022:ACL}
%\citet{Giermindl-etal:2022:EuropJInfSys}
%\citet{Sooraksa:2021:ICEAST}
%\citet{Fernandez:2019}
%\citet{Noack:2019:PsychosocIssHumResMgt}
%\citet{Barrera-etal:2016:LREC}
%\citet{Cristescu-etal:2022:HRM}
%\citet{Uma-etal:2023}
%\citet{Arora-etal:2021:DASA}
%\citet{Melo-etal:2024}

%\citet{Panda-etal:2023:IJIEOM} describe a systematic literature review, bibliometric analysis, and network analysis followed by content analysis around the topic of AI tools in HR can and have improved resilience.

\textbf{Reviews \& Surveys.}
\citet{Rigamonti-etal:2022:SHRMWS} provide a 
recent scoping review of work discussing
HR analytics.
\citet{Tursunbayeva-etal:2023:PersRev}
present a scoping review to
understand how ethical considerations are being discussed by
researchers, industry experts and practitioners, and to identify gaps, priorities and recommendations for
ethical practice.
\citet{Dhameliya-Desai:iPACT} and \citet{Fabris-etal:2023:ArXiv} provide further survey articles that are not subject to the length restrictions of this paper.

\citet{Bartosiak-Modlinski:2023:CarDevInt}
present an empirical study ($N=76$ subjects) to find whether biased suggestions from recommendation systems can cause harm; they find that the indeed can influence individuals who make disciplinary decisions.
\citet{Jotaba-etal:2022:EurJInnovMgt}
present a systematic review of 
innovation and the HR domain in general.

\citet{McCartney-Fu:2023:JOrgEffectPeoplPerf} investigate
to what extent earlier promises of
``HR analytics'' have been realized;
in their quest to explore the state of
the debate, they identify the following
major topics:: inconsistency among the concept and definition of ``people analytics'' (AKA HR analytics),
missing evidence of the impact of people analytics, not being ready to perform people analytics as an organization,
the pople analytics ownership debate,
and ethical/privacy concerns of using people analytics.

\citet{Booyse-Scheepers:2023:MgtResRev}
identify barriers in the adoption of AI for automated organisational decision-making in organizations.

\citet{Johnson-etal:2023:JTourFut} proposed that AI can help with
 scheduling and coordination (chatbots with access to both party's calendars),
computer-assisted interview analysis (attitude towards previous jobs, English proficiency) and  computer-assisted coaching (to improve applicants' interviewing skills and to provide feedback to mentors).

% ----------------------------------------------------------------------

\section{Opportunities}

\begin{figure*}[!h]
    \centering
    \includegraphics[width=.9\textwidth]{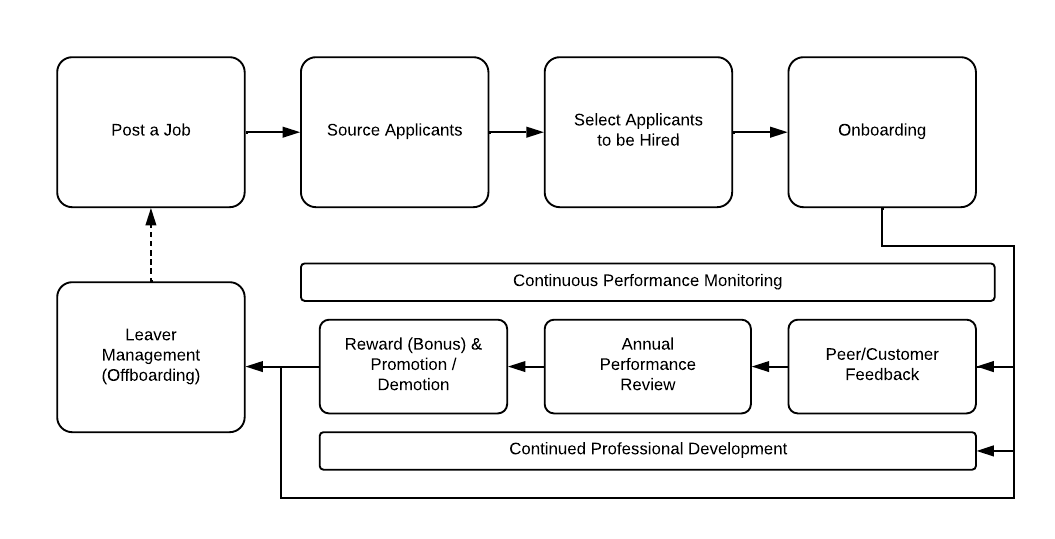}
    \caption{Employee ``Lifecycle''}
    \label{fig:workflow}
\end{figure*}

%% Figure 2 shows the typical employment ``lifecycle'' from posting a job over an employee taking the position until leaving it again eventually. 

Potential employees and hiring managers follow two interconnected workflows, either implicitly or consciously (see Figure \ref{fig:workflow}).
Each stage of this process can arguably benefit from NLP
support: as we have seen in the literature review, assistance
for authoring job postings, choosing the right 
skill descriptors and role titles, ensuring across-gender
suitability on the employer's side already exists and will
no doubt be improved further. On the employee side, enriching
resumes with the right keywords to make a match more likely,
style checkers and format templates are already available, the
next stage is personalized resume crafting assistance that
could even take specific job ads into consideration and
tailor a candidate's presentation to it specifically.
At the Applicant Sourcing stage, better crawlers can find
candidate resumes on personal blogs or homepages so they
may be proactively approached.
At the Applicant Selection stage, matching engines can
ensure that vocabulary mismatch does not lead to false negatives
by using word and document embeddings to encode the
distributional semantics of a resume and a job description.
The On-boarding, Feedback, and Annual Performance Review
stages can be supported by chatbots in the sense that at least
for the evidence gathering phase, it makes sense to provide
automated assistance; we believe that the ultimate review
talks are best kept between two humans; nevertheless it is in
the employer's and in the employee's interest, to conduct
exhaustive evidence of the past year's contributions of the
employee first, and a chatbot could help to avoid forgetting
aspects of the work, the customer relation or relation with
colleagues. Likewise, when an employee leaves, chatbot-type
assistants could help employees leaving an organization to
support running through a check list in order to maximize
compliance, and to avoid spending human time answering
repeat questions.

%% But scale is not the only advantage: the machine will not judge based on skin color (at least not unless programmed to do so), so ethnic minorities could benefit from the machine's radical ``equalitarianism''.

\subsection{Equal Treatment}

While machine learning based methods
(including those in NLP) have been rightly
criticized of bias, at the same time
there is a strong case for automation
from an equality point of view: in the
same way that Justitia, the Roman goddess
of justice, is depicted with her ideas
covered so as to suggest that in front of
the law, all are treated equally, we
know in practice, often people are
treated differently, e.g. based on their
skin color. In contrast, an automated
resume assessment tool 
% will always
can be designed to  behave the same way, regardless whether
the applicant is called ``Dr. John McCarthy''
or  ``Imarogbe Abayomi'' \cite{Bertrand-Mullainathan:2004:AmEconRev}.

\subsection{Low Cost, High Throughput}

Arguably the main opportunity that automation with NLP brings to HR is increase in processing throughput, potentially by orders of magnitude. Automated systems can potentially extract data orders of magnitude faster than would be possible manually. 

The machine causes lower cost when asked to rank a list of human
applicants, about which sufficiently detailed data is already available
in electronic form. It can process a large quantity of resumes in seconds,
it does not require breaks and in this way, results are available in
near real-time.

\subsection{Consistency}

The machine, while in many ways too inflexible and inferior to human
judgment, has one key quality advantage: it acts consistently --
presented with the same evidence, the same model will re-create the
same decision, without variability depending on time of day, hours
worked etc.

\subsection{Selectively Lower Barriers}

While most candidate applicants may prefer to talk to a human 
``flesh and bones'' hiring manager or HR representative,
extremely a subset of introverted, shy or traumatized people may
struggle less inhibited to tell a machine things compared to
talking to humans (though the opposite reaction may
also happen). A software-implemented hiring chatbot may also have
the advantage of not asking any illegal questions (``Are you
married? Do you have children?'').

% ----------------------------------------------------------------------

\section{Challenges and Issues}

\subsection{Lack of Process Transparency}

Applicants outside typically do not know
what happens to their application submitted.
In a perfect world, a recuirter from a recruitment
agency is a proxy of a candidate employee's interests,
but this model does not scale to tens of thousands of
workers.
Recruitment agencies often also act as gatekeeepers
that use -- often keyword-based -- filters
to pre-screen a shortlist, often resulting
in false negatives, i.e. desirable candidates
with poorly written resumes are filtered out.
A potential solution could be a workflow-aware
chatbot that the applicant can ask about the status
(e.g. via email) or that proactively notifies
the applicant about all future stages and the
current state and developments of interest
(``the hiring manager looked at your resume today'').

\subsection{Power Asymmetry and Market Making}

To most people, conceptualizing ``seeking
employment'' as an application process to
one or more potential hiring entities seems
a natural process, especially given the
size and power of the typical corporation;
however, an alternative possible
scenario could be imagined where employees
auction their workforce to the highest
bidding company instead, thus letting
companies/employers/hiring managers compete
for their talent. Such a scenario is as
asymmetric as the current one, and a more
symmetric alternative would be using methods
somewhat akin to the one devised by
the Nobel prize winning ``stable marriage''
algorithm for bi-partite matching
by Alvin E. Roth and Lloyd S. Shapley \cite{Gale-Shapley:1962:AmMathMon,Roth:1984:JPolitEcon}, with the exception that seeking employment, unlike a good marriage, is a dynamic, repeated version of the game.\footnote{\url{https://www.nobelprize.org/prizes/economic-sciences/2012/summary/} (accessed 2023-12-08)}

\subsection{Lack of Privacy and Conflict of Interest}

Conflict of interest exists, for instance, when the
platform where people find out about open positions and apply
for them is operated by the same corporation that one applies;
for example, Microsoft owns LinkedIn (which we note is not
Microsoft-branded), which is a major player in the job ad posting, dissemination, search, application and matching space,
but Microsoft is also a huge technology company that LinkedIn
users apply a job for. As Microsoft owns the platform, it also
knows where else an applicant has applications going on.
Not only that, Microsoft and Google also operate email services,
Outlook and GMail, so they have in principle access to all
email conversations with recruiters, companies and friends.
Google Hangout and Skype (also owned by Microsoft) are platforms
commonly used for remote job interviews.

\subsection{Mis-characterization of the Applicant}

There is a risk that job applicants are mis-characterized because
their resumes do not use the language that the recruitment
platform expects to match them against a desired job description.
For example, if the matching engines is crude and keyword-based,
and the resume mentions a passion for functional programming,
Scheme and Haskell, whereas the job description demands strong
LISP development skills, encoded as a mandatory match of the
keyword ``\texttt{LISP}'', the problem ensues that the matching
engine fails to have knowledge that Scheme is a flavour of the
LISP language. As a consequence, potentially strong candidates
would be ignored, resulting in a false negative problems.

\subsection{De-Humanizing the Applicant}

While it may be beneficial from a discrimination point of view
not to have photos and names on resumes (for instance, to avoid
racial bias), the automation of the processes of hiring and HR management by using more NLP may eventually also eventually suffer from an element of de-humanizing the applicant. Human contact between potential future colleagues and actual colleagues remains a desirable element of the recruiting
process and the process of people management (arguably, using the term ``human resource'' itself abstracts people to mere ``resources'').

\subsection{Disadvantaging Underresourced Languages}

Because NLP components are most mature for the English language,
by a large margin less so for all other languages, and not
available for so-called ``low density'' (AKA ``lowres or ``underresourced languages''), candidates with resumes written
in such languages will be disadvantaged in the international
job market, and processes will be less efficient in those
countries where mostly underresourced languages are spoken and
written as the main means of communication.

\subsection{Lack of Diversity}

Because NLP components may be based on trainable machine learning models,
the present-day workforce is the most likely training set; however, this
leads to ``more of the same'' hiring, which is what anecdotally happened
at Amazon \cite{Dastin:2018}, where non-males were rejected because in the training
data, all high performers were male (simply because it was mostly male
generally speaking, regardless of performance).

% ----------------------------------------------------------------------

\section{Summaries, Conclusions and Future Work}

In this paper, we have identified, collected, analyzed and presented
potential challenges/issues and chances/opportunities pertaining to the application of language technology
to the field of hiring and human resource management.

We can now revisit our initial research questions:\\
\textit{1. What types of support can NLP enable?} Most hiring
and HRM workflows can benefit from NLP-enabled support: along
the employee lifecycle, we have identified a rich range of
potententially valuable functions where NLP can play a role. \\
\textit{2. How should the support differ from the employer's side versus the employee's side?} Each of the two perspectives,
the employees and the employers, deserve their respective
software stack that helps them to maximize their opportunities.
Employees want to be found, look good (generally employable and
match to specific job postings), whereas employers want to
probe the claimed qualifications, experiences. Where both
perspectives are actually aligned is that matching the
most suitable job which the most candidate is likely
best for both.\\
\textit{3. What opportunities versus challenges/risks result from the recent technological advances?} Faster and better recruiting
experience for both sides are perhaps the main upsides, and
the main risks are that less IT-savvy candidate employees may
be left behind.

%In terms of numbers, the issues prevail, but this may be misleading as they do not all carry the same weight.
%
In future work, ways to address the issues pointed out here
should be developed and implemented. Where this may prove difficult or impossible, laws and regulations should be introduced to protect human employees.

% -----------------------------------------------------

% \clearpage

% \bibliographystyle{splncs04}
\bibliography{Leidner-Stevenson-2024-NLP4HR}

\nocite{*}

\end{document}